\documentclass[runningheads]{llncs}
\usepackage{graphicx}
\usepackage{amssymb}
\usepackage{amsmath}
\usepackage{orcidlink}
\begin{document}
\title{Effectiveness of L2 Regularization in Privacy-Preserving Machine Learning\thanks{This work was partially supported by the European Commission (EC) through the European Union’s Horizon Europe Framework Programme within the frame and for the purpose of the artificial Intelligence To enHAnce Civic pArticipation (ITHACA) project under Grant Agreement No. 101094364.}}
\author{Nikolaos Chandrinos\orcidlink{0009-0004-0737-3261}\inst{1} \and
Iliana Loi\orcidlink{0000-0001-9112-0638}\inst{2}
\and
Panagiotis Zachos\orcidlink{0000-0003-0460-252X}\inst{2}
\and
Ioannis Symeonidis\orcidlink{0000-0002-5050-8617}\inst{1} \and
Aristotelis Spiliotis\orcidlink{0000-0001-9042-2044}\inst{1} \and
Maria Panou\orcidlink{0000-0002-2161-9246}\inst{1} \and
Konstantinos Moustakas\orcidlink{0000-0001-7617-227X}\inst{2}
}
\authorrunning{N. Chandrinos et al.}
\institute{Human Factors and Vehicle Technology, Hellenic Institute of Transport, Centre for Research and Technology Hellas, Thermi, Greece\\
\email{\{chandrin, ioannis.sym, aspiliotis, mpanou\}@certh.gr} \and
Wire Communications and Information Technology Laboratory, Dept. of Electrical and Computer Engineering, University of Patras, Patras, Greece\\
\email{loi@ceid.upatras, p\_zachos@ac.upatras.gr, moustakas@ece.upatras.gr}
}
\maketitle              % typeset the header of the contribution
\begin{abstract}
Artificial intelligence, machine learning, and deep learning as a service have become the status quo for many industries, leading to the widespread deployment of models that handle sensitive data. Well-performing models, the industry seeks, usually rely on a large volume of training data. However, the use of such data raises serious privacy concerns due to the potential risks of leaks of highly sensitive information. One prominent threat is the Membership Inference Attack, where adversaries attempt to deduce whether a specific data point was used in a model's training process. An adversary's ability to determine an individual's presence represents a significant privacy threat, especially when related to a group of users sharing sensitive information. Hence, well-designed privacy-preserving machine learning solutions are critically needed in the industry. In this work, we compare the effectiveness of L2 regularization and differential privacy in mitigating Membership Inference Attack risks. Even though regularization techniques like L2 regularization are commonly employed to reduce overfitting, a condition that enhances the effectiveness of Membership Inference Attacks, their impact on mitigating these attacks has not been systematically explored.
 
\end{abstract}
\section{Introduction} \label{sec:intro}
Privacy, a multifaceted concept encompassing freedom of thought, control over personal information, and protection from surveillance, has become increasingly complex in the digital age. As industries widely adopt Artificial Intelligence (AI) and Machine Learning (ML) technologies, they rely on vast amounts of sensitive data to train sophisticated models. This reliance raises significant concerns about data privacy, especially with the advent of Machine Learning as a Service (MLaaS) platforms and important regulations like the General Data Protection Regulation (GDPR). Addressing these challenges, this paper explores how Privacy-Preserving Machine Learning (PPML) techniques can safeguard sensitive information. Specifically, we investigate the effectiveness of $L_2$ regularization in enhancing defenses against Membership Inference Attacks (MIAs), comparing its performance to commonly used methods such as Differential Privacy (DP).

Privacy, in general, is a concept of disarray. A broad term that accumulates, freedom of thought, control over one’s body, seclusion in one’s home, control over personal information, freedom from surveillance, protection of one’s reputation, and protection from searches and interrogations, among other things \cite{solove2010understanding,xu2021privacy}. A similar difficulty exists for privacy in the digital realm. Typically digital privacy emphasizes protecting personal identity involving ensuring individuals' control over their personal information and safeguarding against surveillance \cite{xu2021privacy}.

PPML plays a crucial role in upholding these standards. PPML emerged to address privacy challenges posed by ML and Deep Learning (DL) applications \cite{tanuwidjaja2020privacy,xu2021privacy}. PPML techniques aim to develop methodologies to train and deploy models without compromising sensitive personal data. This includes methods like encryption, anonymization, DP, etc., which protect data during both the training phase and when making inferences.

Despite these advancements, several types of cybersecurity threats stake the integrity of sensitive data used in models. These threats can be broadly categorized into three types \cite{liu2021machine}: poisoning attacks, where adversaries manipulate the training data to corrupt the learning process; evasion or exploratory attacks, which aim to deceive the model into making incorrect predictions; and inference attacks, where the goal is to extract information about the data or the model itself rather than influencing its output \cite{xu2021privacy}.

Regarding inference attacks, common examples include Membership Inference Attacks (MIAs), which we'll discuss in this work, in which an attacker can infer whether a data point was used to train the model \cite{song2021systematic,xu2021privacy,shokri2017membership}. This black-box access to predictions can estimate aspects, whether a specific patient profile was used in the training dataset \cite{xu2021privacy,kaya2020effectiveness,ying2020privacy}.

The purpose of this paper is not to provide a comprehensive tutorial on the tools of PPML per se but rather to shed light on how $L_2$ regularization can further enhance mitigation against MIAs compared to common practices like DP. The rest of this paper is structured as follows. Section~\ref{sec:preliminaries} provides the necessary background, while the experimental evaluation is provided in Section~\ref{sec:experiments}. Finally, conclusions are drawn in Section~\ref{sec:conclusions}.

\section{Preliminaries} \label{sec:preliminaries}
\subsection{Deep Neural Networks}
Deep Neural Networks (DNNs) are composed of multiple layers of interconnected neurons, where each layer transforms its input data, often into a higher-dimensional representation \cite{DL1,DL2}. The operation of a fully connected (dense) layer can be mathematically described by:
\[
    \mathbf{y} = \phi\left(\mathbf{W}\mathbf{x}+\mathbf{b}\right),
\]
where: $\mathbf{x}\in\mathbb{R}^{m}$ in the input vector, $\mathbf{W}\in\mathbb{R}^{n \times m}$ is the weight matrix, $\mathbf{b}\in\mathbb{R}^{n}$ is the bias vector, $\phi(\cdot)$ is a non linera activation function (e.g. ReLU, sigmoid), $\mathbf{y}\in\mathbb{R}^{n}$ is the output vector \cite{DL1,DL2,DL3}.

The strength of DNNs lies in their capacity to model complex nonlinear relationships throughout their architecture. This ability enables the network to learn hierarchical feature representations from the data \cite{DL1}. A DNN with $L$ layers can be represented as:
\[
\mathbf{y} = f\left(\mathbf{x};\Theta\right) = f^{(L)}\left(f^{(L-1)}\left(\dots f^{(1)}(\mathbf{x}\right)\right),
\]
where: $f^{(l)}(\cdot)$ denotes the function computed at layer $l$, $\Theta = \left\{\mathbf{W}^{l},\mathbf{b}^{(l)}\right\}^{L}_{l=1}$ represents all the networks parameters \cite{DL1}.

Training a DNN involves finding the optimal set of parameters $\Theta$ that minimize a predefined loss function $\mathcal{L}\left(\mathbf{y},\mathbf{\hat{y}}\right)$, where $\mathbf{\hat{y}}$ is the true label corresponding to the input sample $\mathbf{x}$. The loss function measures the discrepancy between the predicted output $\mathbf{y}$ and the actual output $\mathbf{\hat{y}}$. For multi-class classification problems, the cross-entropy loss function is frequently employed \cite{DL1}.

The optimization of the loss function is typically carried out using gradient-based methods. The most common approach is Stochastic Gradient Descent (SGD) and its variants, such as Adam and RMSprop. The gradients of the loss with respect to the network parameters are computed using the backpropagation algorithm. Backpropagation applies the chain rule from calculus to efficiently propagate errors backward through the network layers, enabling the adjustment of weights and biases to minimize the loss \cite{DL1}.

\subsection{L2 Regularization} \label{sec:L2}

$L_2$ regularization, also known as Ridge regression, is a widely utilized technique in machine learning, particularly in scenarios where model complexity needs to be controlled without eliminating features. The core idea behind $L_2$ regularization is to penalize large coefficients in the model, thereby reducing the risk of overfitting \cite{shalev,bickel2006regularization}.

In machine learning, $L_2$ regularization modifies the loss function by adding a penalty to the sum of the squared coefficients of the model parameters. The modified loss function can be expressed as:

\[
\text{Loss} = \text{Loss}_\text{orig} + \lambda \sum_{i=1}^{n} w_i^2
\]

Here, $\text{Loss}_\text{orig}$ represents the original loss function (such as mean squared error for linear regression), $\lambda$ is the regularization parameter that controls the strength of the penalty, and $w_i$ are the coefficients of the model. The term $\sum_{i=1}^{n} w_i^2$ is the $L_2$ penalty, which increases as the magnitude of the coefficients increases \cite{shalev}.

The effect of this penalty is to shrink the coefficients towards zero, but not necessarily to zero. This shrinkage prevents any single feature from dominating the model, especially in the presence of multicollinearity, where features are highly correlated \cite{chan2022mitigating}. The regularization parameter $\lambda$ plays a crucial role; a higher $\lambda$ value imposes a stronger penalty, leading to greater shrinkage of the coefficients \cite{shalev}.

In neural networks, $L_2$ regularization is incorporated into the training process through a technique known as weight decay. The loss function, often based on cross-entropy or mean squared error, is augmented with an $L_2$ penalty on the network’s weights. The modified loss function in this context can be written as:

\[
\text{Loss} = \text{Loss}_\text{orig} + \lambda \sum_{j=1}^{m} \sum_{k=1}^{n} w_{jk}^2
\]

Where $w_{jk}$ represents the weights connecting the neurons between layers $j$ and $k$, $m$ is the number of layers, and $n$ is the number of connections \cite{shalev}.

The inclusion of the $L_2$ penalty ensures that the network learns smaller weight values, which helps in preventing the model from overfitting to the training data. During the training process, the gradient update rule for the weights is modified to include this penalty. Specifically, the weights are updated according to:

\[
w_{jk} \leftarrow w_{jk} - \eta \left( \frac{\partial \text{Loss}_\text{orig}}{\partial w_{jk}} + 2\lambda w_{jk} \right)
\]

Where $\eta$ is the learning rate, and the term $2\lambda w_{jk}$ represents the derivative of the $L_2$ penalty. This update rule gradually reduces the magnitude of the weights as the training progresses, leading to a simpler and more generalizable model \cite{shalev}.

The use of $L_2$ regularization in neural networks improves the model’s ability to generalize, reduces sensitivity to input noise, and enhances the stability of the learning process \cite{shalev}. However, the choice of the regularization parameter $\lambda$ is critical; if set too high, it can lead to underfitting, where the model is overly simplistic and fails to capture the underlying data patterns \cite{shalev,bickel2006regularization}.

\subsection{Memberships Inference Attack} \label{sec:MIA}
MIAs originate from the observation that models usually perform differently on the data they are trained on than first-time-seen data. Generally, it refers to acquiring knowledge about whatever a given data sample $(\overrightarrow{x^*}, y^*) \in D$, where $D$ is the training dataset used to train a model $f$. $D$ contains samples used for training. $D'$, the testing dataset, has an almost identical distribution as the records in $D$, however, they never participated in training $f$.  This deference means that an instance $\overrightarrow{x^*} \in D$ is a member, while $\overrightarrow{x'^*} \in D'$ is a non-member. Therefore, MIAs are used to determine whether a given record belongs to the training dataset or not \cite{hu2021ear,tanuwidjaja2020privacy}. 

In certain situations, identifying that a person is included in the training set can have privacy implications, especially in a medical study involving patients with a rare disease. Additionally, membership inference can be used as a foundation for launching data extraction attacks. \cite{tanuwidjaja2020privacy,vassilev2024adversarial}.

MIAs may extend to extracting specific characteristics of sensitive training data or potentially reconstructing it entirely. Typically, such attacks exploit the prevalent issue of overfitting, which leads to a noticeable accuracy gap between the training data and external datasets \cite{tanuwidjaja2020privacy,liu2021machine}.

The Attacker Advantage (AA) metric provides a measure on how much additional information the attacker gains with each additional iteration by calculating the difference between the probability of the adversary correctly guessing a data point was included in the training set and the probability of the adversary correctly guessing a data point was not included in the training set \cite{yeom2018privacy}.

The AA is evaluated as follows.  Let $\mathcal{A}$ be an adversary, A be a learning algorithm, n be a positive integer, and D be a distribution over data points (x,y).
The membership experiment is developed as follows

\begin{itemize}
    \item Sample $S \sim D^n$, and let $A_S = A(S)$.
    \item Choose $b \leftarrow {0,1}$ uniformly at random.
    \item Draw $z \sim S$ if $b=0$, or $z\sim D$ if $b=1$.
    \item $Exp^M(\mathcal{A},A,n,\mathcal{D})$ is 1 if $\mathcal{A}(z,A_S,n,\mathcal{D})=b$ and 0 otherwise. $\mathcal{A}$ must output either 0 or 1.
\end{itemize}

The membership advantage of $\mathcal{A}$ is defined as
\[
    \mathsf{Adv}^\mathsf{M}(\mathcal{A},A,n,D)=2\Pr[\mathsf{Exp}^\mathsf{M}(\mathcal{A},A,n,D)=1]-1,
\]
where the probabilities are taken over the coin flips of $\mathcal{A}$, the random choices of S and b, and the random data point $z\sim S$ or $z\sim\mathcal{D}$.

Equivalently, the right-hand side can be expressed as the difference between $\mathcal{A}$'s true and false positive rates
\[
    \mathsf{Adv^M}\Pr[\mathcal{A}=0 | b=0] - \Pr[\mathcal{A}=0 | b=1],
\]
where $\mathsf{Adv^M}$ is shorthand for $\mathsf{Adv}^\mathsf{M}(\mathcal{A},A,n,D)$.

\subsection{Differential Privacy} \label{sec:DP}
Differential privacy (DP) techniques apply arbitrary modifications to a dataset such that if an individual has access to the dataset’s entries, they will not be able to infer any personalized sensitive information from it \cite{xu2021privacy,Zapechnikov2020}. Such methods include the addition of random noise to the data through differential procedures (Gaussian, Laplace, exponential, etc.).

Some of the most recent works on differential privacy preservation are \cite{Abadi2016,Phan2017,Arachchige2019,Phan2020}. One of the most prominent works in PPML is \cite{Abadi2016}, where Gaussian noise is injected to the gradients of a neural network’s parameters at each training step, rendering the model capable of learning the so-called differentially private parameters. Similarly, in \cite{Phan2017}, an adaptive Laplace algorithm named AdLM was developed to also learn differentially private parameters, but with adaptively adding noise to input features of a neural model according to their relevance to the network’s output. Explanatory, the main idea behind AdLM is to add less Laplace noise to the features that contribute the most to the model’s output, while adding more noise to those that are less relevant. Furthermore, noise is added adaptively into affine feature transformations and loss functions as well. In contrast to \cite{Abadi2016}, in this approach the added noise and privacy consumption do not accumulate in each epoch, since AdLM does not access the ‘clean/unmasked’ data during the training process, rendering AdLM independent of the number of training epochs. In other words, AdLM adds Laplacian noise as a preprocessing procedure rather than on training time. Another benefit of this framework is that it has the ability to be incorporated into a plethora of deep learning models. The same team of authors in one of their newest works \cite{Phan2020}, propose a scalable framework, named StoBatch, to preserve differential privacy in deep adversarial learning. StoBatch is a stochastic batch mechanism that maintains the privacy in the learning of model parameters by first introducing noise to both input features and their latent space and then integrating adversarial learning to enhance the decision bounds of the model. In contrast to \cite{Phan2017}, the amount of noise introduced to the model is minimized by incorporating an adversarial objective function that combines the loss function for training data with a loss function for differentially private adversarial data, hence preserving the model’s utility (i.e., preventing privacy data and noise accumulation over training). Moreover, by feeding both input and hidden layers of a model with noise renders StoBatch more resistant and robust against adversarial data. The stochastic batch training that StoBatch utilizes, separates data in local trainers to learn differential privacy parameters and then the computed gradients from the trainers are combined, allowing the calculation of adversarial data from various data batches at each epoch. The latter enables StoBatch to be scalable to vast datasets and deep learning models \cite{Phan2020}.

The aforementioned approaches use global differential privacy, which presupposes that both the data and differential privacy method to apply random noise to these data, reside in a server. Nevertheless, local differential privacy enables data owners to ‘hide’ any sensitive/personal data locally before sharing them. Hence, local differential privacy is considered to be a more secure way to privacy-proof training data as well as computationally cheap, since it does not require a vast number of computational resources like global differential privacy \cite{Arachchige2019}. A local differentially private method consists of LATENT \cite{Arachchige2019}, which acts as an intermediate layer into a deep learning model that a data owner can utilize to randomize their data before distributing them to untrusted users/neural networks. LATENT was evaluated by being integrated between the convolutional and the fully connected layers of a CNN model. In particular, LATENT randomizes the 1D real-value flattened vectors (corresponding to an image input) that the convolutional layers produce by transforming them to discrete vectors \cite{Arachchige2019}.

It is worth mentioning the Tensorflow Privacy \cite{tfprivacy}, an open-source Python library by Google Research, which enables the creation and training of privacy-preserved machine learning models. Especially, Tensorflow Privacy utilizes Tensorflow \cite{tfprivacy} optimizers for training an ML model using differential privacy algorithms, while it is able to compare neural networks in privacy measures and preserve model utility. It is worth mentioning that Tensorflow is a widely used open-source Python package that supports the development of ML/DL models, thus, not supporting PPML creation. Therefore, the Tensorflow Privacy framework’s main aim is to facilitate the adoption of privacy-preserving methods into ML using Tensorflow or other APIs such as Keras.

\section{Experimental Results} \label{sec:experiments}
In this work, we experimentally evaluate the MIAs performance in two different datasets: (a) MNIST \cite{mnist}, (b) CIFAR10 \cite{cifar10}, by applying two neural network architectures: (a) a fully connected neural network (FCNN), (b) a convolutional neural network (CNN) and (c) an augmented version of the Toxic Tweets Dataset for text classification \cite{kaggleToxicTweets}, covering a wide range of possible scenarios with commonly used architecture designs. For each aforementioned scenario, we compared two training methods: (a) a normal (baseline) approach using a non-DP-trained model, and (b) a DP method involving a DP-trained model. Additionally, for the text classification task, we use a simple embedding-based neural network tailored for natural language data. To study how $L_2$ regularization affects their privacy, we also evaluated both methods with $L_2$ regularization. Finally, we extend our experimental evaluation to examine how the difference in accuracy between training and evaluation affects MIAs.

\subsection{Image Classification Task}

The MNIST \cite{mnist} dataset consists of handwritten digits, including 60,000 samples in the training set and 10,000 in the evaluation set. The input flattened images (784 features) are fed to a fully connected network which consists of three layers with 10, 20, and 10 neurons, respectively. The non-DP models are optimized for 50 epochs using Adam optimizer \cite{adam} with a learning rate equal to 0.0001 and using the cross-entropy loss. Batches of 32 samples were used. A similar configuration was used for the DP models, but instead of Adam, we used DP-Adam of TensorFlow Privacy \cite{tfprivacy}.

\begin{figure}[ht!]
\includegraphics[width=\textwidth]{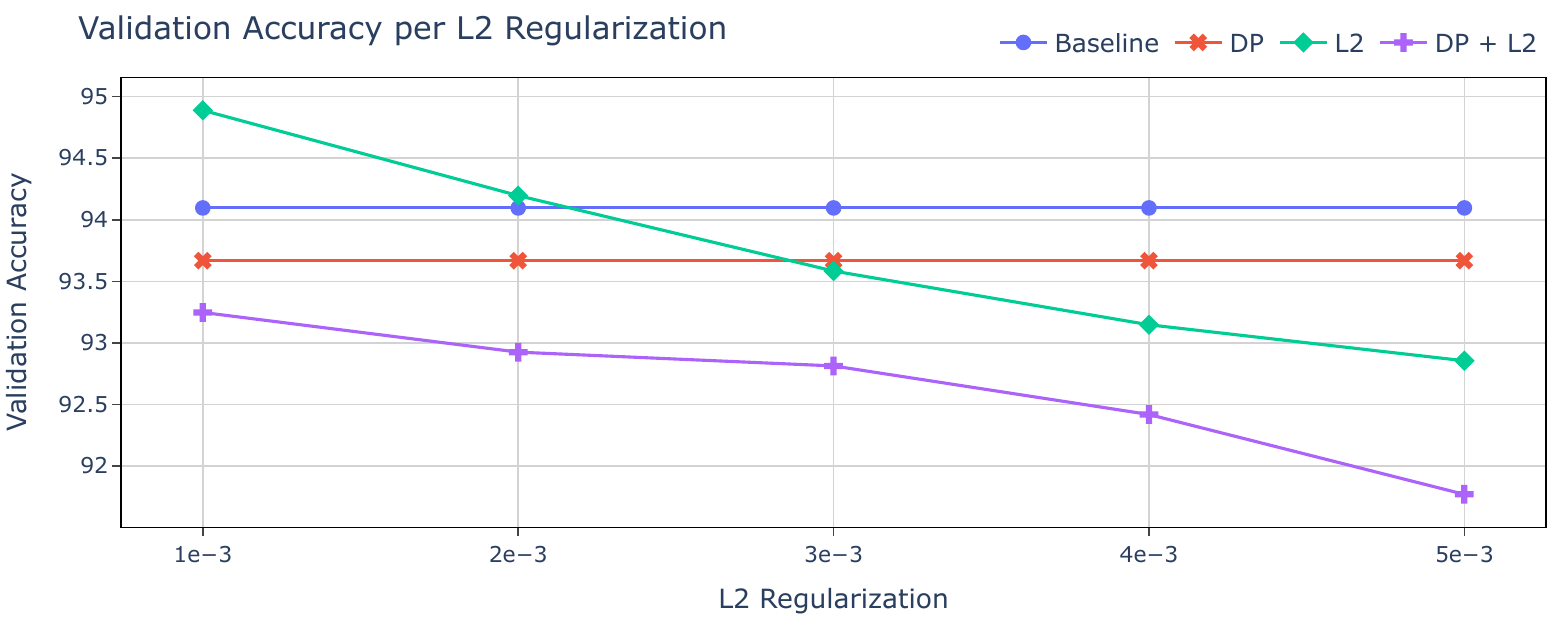}\\
\includegraphics[width=\textwidth]{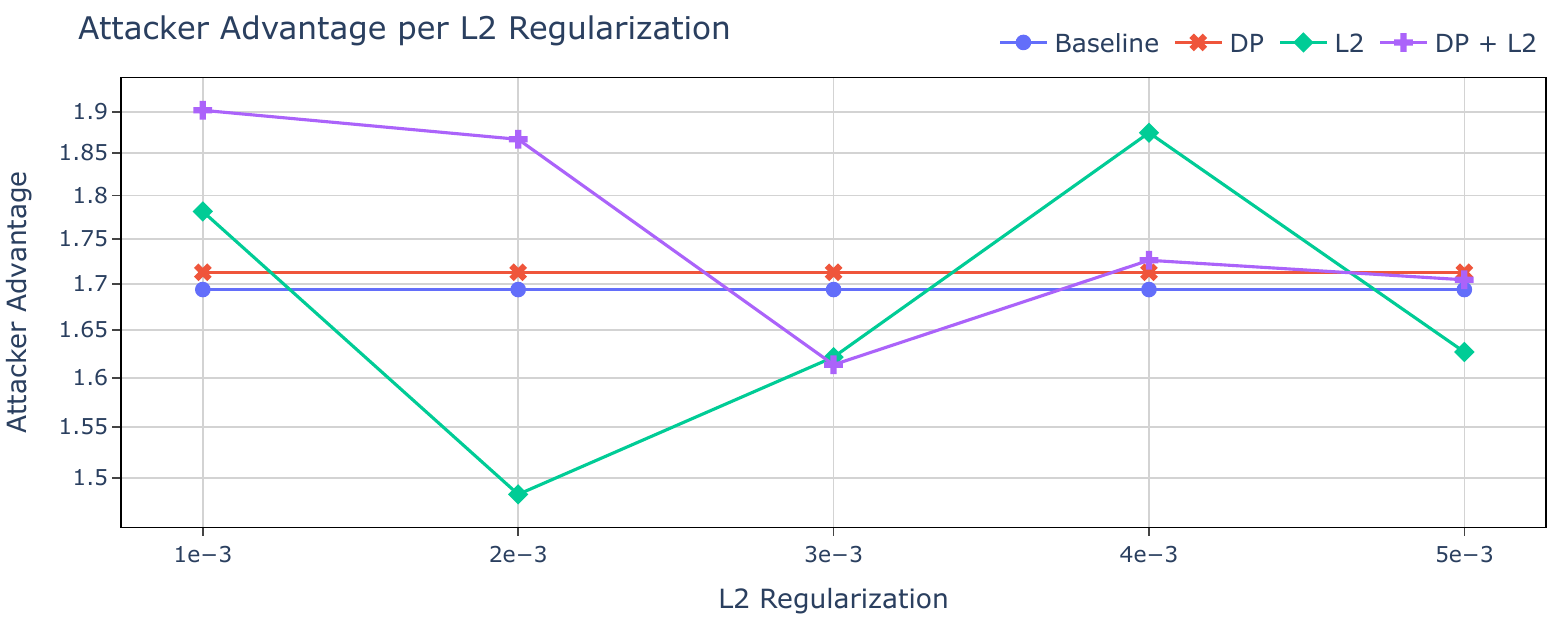}
\caption{$L_2$ regularization on the FCNN model trained on the MNIST dataset. The figure shows how validation accuracy and attacker advantage change with different $L_2$ regularization strengths ($\lambda$) for both Baseline (non-DP) and DP models.} \label{fig:mnist_l2}
\end{figure}

The experimental results are reported in Table \ref{table:mnist_l2} and visualized in Figure \ref{fig:mnist_l2}, where we report the average and variance of the training, evaluation accuracy, and attacker advantage over five training runs for each evaluated method. Furthermore, we demonstrate the performance of every method on different $\lambda$ values with $L_2$ regularization.

\begin{table}[ht!]
\caption{Performance metrics for FCNN on the MNIST dataset with varying $L_2$ regularization strengths ($\lambda$). The table reports the average and standard deviation of training accuracy, validation accuracy, and attacker advantage over five runs for both Baseline (non-DP) and DP models.}\label{table:mnist_l2}
\centering
\small
\setlength{\tabcolsep}{3pt}
\begin{tabular}{l|cccccc}
\hline
\hline
 & \multicolumn{6}{|c}{Train Accuracy} \\ 
\hline
\hline
Baseline & $94.6 \pm .250$ & $94.8 \pm .171$ & $94.0 \pm .201$ & $93.4 \pm .125$ & $93.0 \pm .144$ & $92.4 \pm .223$\\
DP       & $94.1 \pm .232$ & $93.3 \pm .471$ & $93.0 \pm .318$ & $92.8 \pm .190$ & $92.4 \pm .407$ & $91.7 \pm .153$\\
\hline
\hline
 & \multicolumn{6}{|c}{Validation Accuracy} \\
\hline
\hline
Baseline & $94.0 \pm .361$ & $94.8 \pm .273$ & $94.1 \pm .252$ & $93.5 \pm .266$ & $93.1 \pm .200$ & $92.8 \pm .350$\\
DP       & $93.6 \pm .327$ & $93.2 \pm .415$ & $92.9 \pm .423$ & $92.8 \pm .218$ & $92.4 \pm .330$ & $91.7 \pm .193$\\
\hline
\hline
 & \multicolumn{6}{|c}{Attacker Advantage} \\
\hline
\hline
Baseline & $1.69 \pm .156$ & $1.78 \pm .615$ & $1.48 \pm .615$ & $1.62 \pm .216$ & $1.87 \pm .319$ & $1.62 \pm .182$\\
DP       & $1.71 \pm .340$ & $1.90 \pm .214$ & $1.86 \pm .212$ & $1.61 \pm .135$ & $1.72 \pm .302$ & $1.70 \pm .191$\\
\hline
\hline
$\lambda$ & $0$ & $.001$ & $.002$ & $.003$ & $.004$ & $.005$\\
\hline
\end{tabular}
\end{table}

From Table \ref{fig:mnist_l2} we can observe, that the non-DP model achieves higher validation accuracy compared to DP. Specifically, when the non-DP model is trained with $L_2$ regularization at strengths of 0.001 and 0.002, it attains higher validation accuracy. This shows how mild $L_2$ regularization helps in optimizing the generalization of the model without introducing significant noise into the training process, as is the case with DP.

For models utilizing DP, the introduction of $L_2$ regularization appears to further reduce validation accuracy. This decrement is likely due to the effect of noise addition from both DP and the constraints imposed by $L_2$ regularization, which may restrict the model's capacity to fit the training data effectively. The downward trend in accuracy with increasing $L_2$ regularization strengths across both DP and non-DP models could be attributed to over-regularization \cite{thoma2017analysis}.

Concerning privacy performance measured by attacker advantage, the models exhibit a relatively stable trend across different configurations. This stability suggests that neither the pressure from $L_2$ regularization nor the noise from DP significantly alters the fundamental privacy characteristics of the models in the context of the MNIST dataset. The small difference in train and validation accuracy, presented in Table \ref{table:mnist_l2}, across all models also indicates that overfitting is likely not a prevailing issue in this scenario. The similar attacker advantage across varying configurations supports, that in inference all models maintain a similar level of robustness against such privacy threats.

\subsubsection{The CIFAR10} dataset includes 50,000 images in the training set and 10,000 in the evaluation set with 32 × 32 color image samples that contain one of the 10 object classes. The applied CNN consists of three convolutional layers followed by a single fully connected layer. In more detail, three convolutional layers with 32, 64, 128 filters of size 3 × 3 each of them followed by an average pooling layer 2 × 2. Finally, the extracted features are flattened and fed to the classification layer of 10 neurons. The non-DP CNN is optimized for 50 epochs using the Adam optimizer \cite{adam} with a learning 0.0001, while batches of 32 are used. For the DP method, only the Adam optimizer was changed with the equivalent DP-Adam optimizer of Tensorflow Privacy \cite{tfprivacy}.

\begin{table}[ht!]
\caption{Performance metrics for CNN on the CIFAR10 dataset with varying $L_2$ regularization strengths ($\lambda$). The table presents the average and standard deviation of training accuracy, validation accuracy, and attacker advantage over five runs for both Baseline (non-DP) and DP models.}\label{table:cifar10_l2}
\centering
\small
\setlength{\tabcolsep}{3pt}
\begin{tabular}{l|cccccc}
\hline
\hline
 & \multicolumn{6}{|c}{Train Accuracy} \\ 
\hline
\hline
Baseline & $81.2 \pm .444$ & $77.6 \pm .410$ & $72.7 \pm .345$ & $69.3 \pm .364$ & $66.5 \pm .356$ & $64.6 \pm .342$\\
DP       & $61.4 \pm .303$ & $59.6 \pm .798$ & $59.7 \pm .470$ & $59.9 \pm .347$ & $59.0 \pm .332$ & $59.1 \pm .661$\\
\hline
\hline
 & \multicolumn{6}{|c}{Validation Accuracy} \\
\hline
\hline
Baseline & $72.6 \pm .244$ & $73.3 \pm .358$ & $70.2 \pm 2.23$ & $67.8 \pm 1.23$ & $66.0 \pm 1.08$ & $64.0 \pm .900$\\
DP       & $59.7 \pm .257$ & $58.3 \pm .651$ & $58.5 \pm .985$ & $58.9 \pm .818$ & $58.2 \pm .707$ & $57.9 \pm .647$\\
\hline
\hline
 & \multicolumn{6}{|c}{Attacker Advantage} \\
\hline
\hline
Baseline & $9.25 \pm .298$ & $5.76 \pm .450$ & $3.26 \pm .330$ & $2.05 \pm .251$ & $1.75 \pm .159$ & $1.21 \pm .267$\\
DP       & $2.01 \pm .153$ & $1.75 \pm .098$ & $1.70 \pm .221$ & $1.46 \pm .319$ & $1.45 \pm .165$ & $1.58 \pm .208$\\
\hline
\hline
$\lambda$ & $0$ & $.001$ & $.002$ & $.003$ & $.004$ & $.005$\\
\hline
\end{tabular}
\end{table}

In Table \ref{table:cifar10_l2}, we report the average values and variance of the training and evaluation accuracy, as well as the attacker advantage, across five training runs. Unlike the MNIST case, for this more challenging CIFAR10 dataset, we observe a higher difference between training and evaluation accuracy, which also corresponds to higher values of the attacker advantage. This highlights the importance of incorporating methods to improve privacy, as Table \ref{table:cifar10_l2} demonstrates how higher accuracy differences lead to an increase in the attacker’s advantage. This phenomenon will be further investigated, as it suggests that models with larger training-to-evaluation accuracy differences tend to have higher vulnerability to attacks.

\begin{figure}[ht!]
\centering
\includegraphics[width=\textwidth]{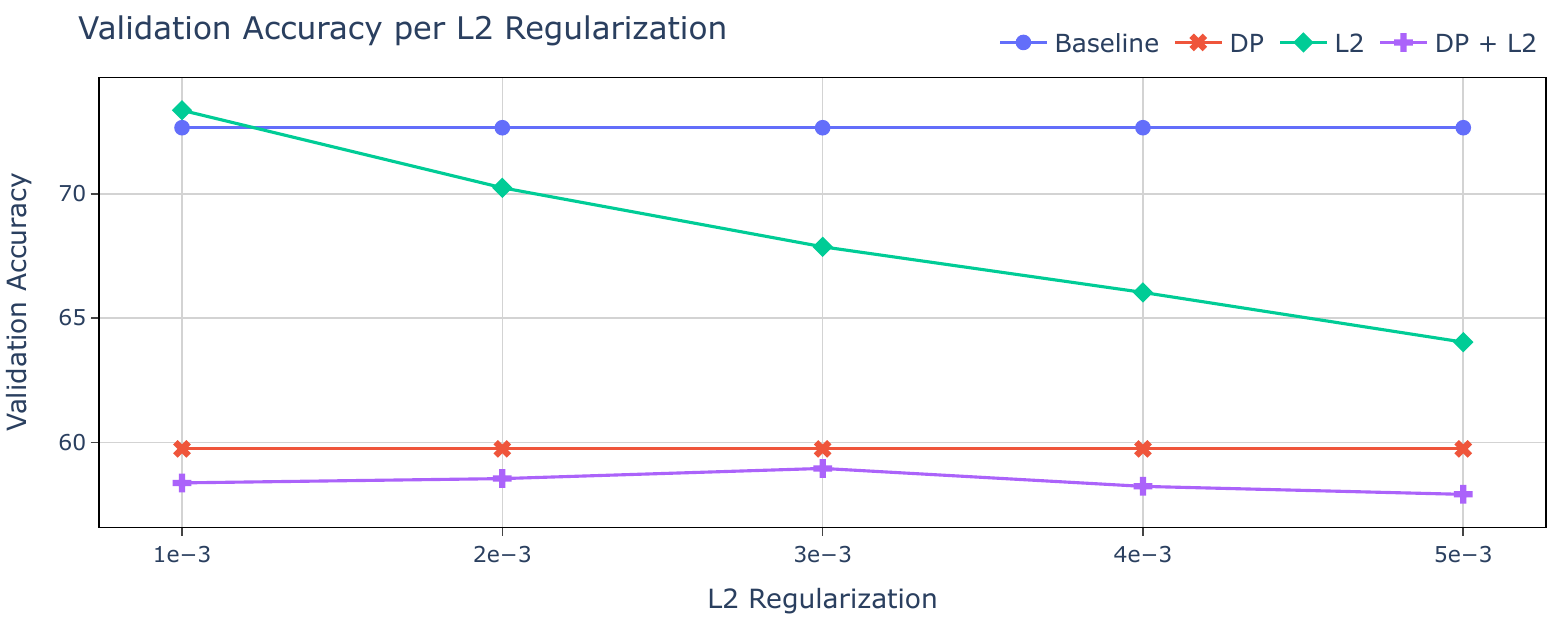}\\
\includegraphics[width=\textwidth]{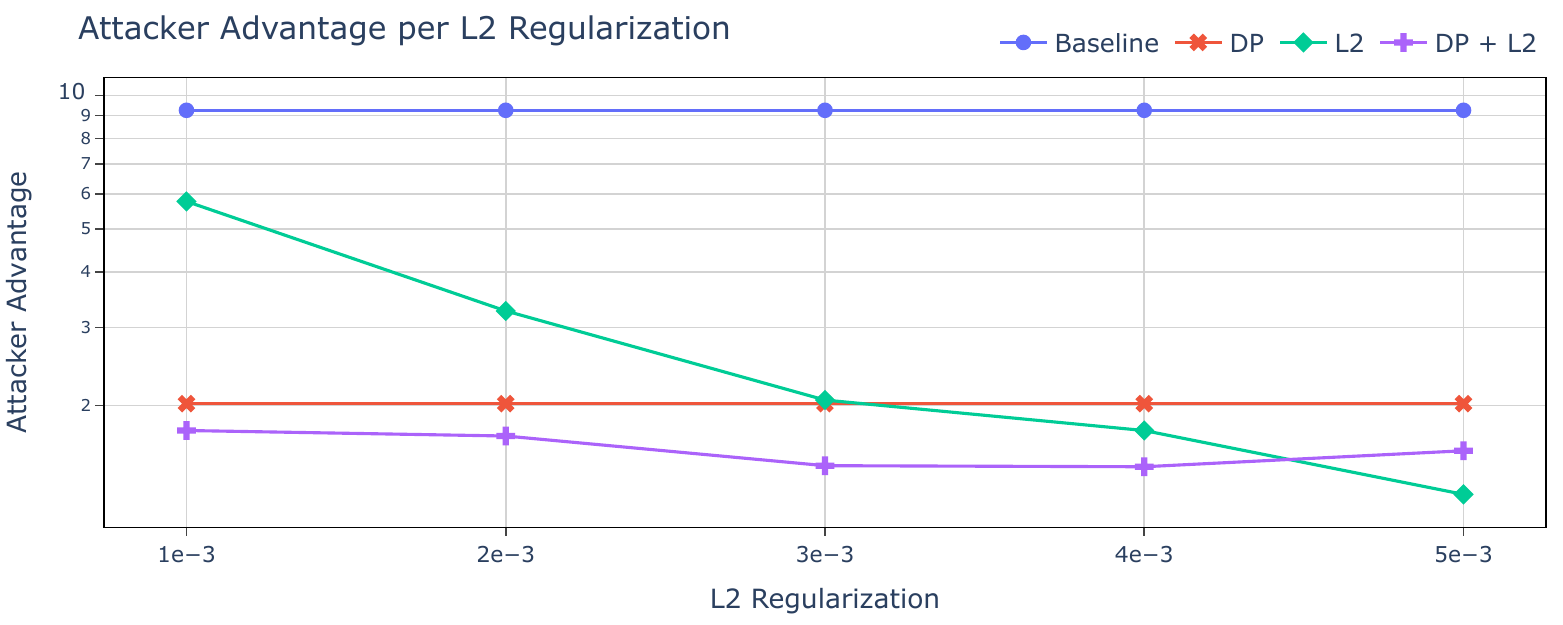}
\caption{Effect of $L_2$ regularization on the CNN model trained on the CIFAR10 dataset. The figure illustrates the relationship between validation accuracy and attacker advantage across different $L_2$ regularization strengths ($\lambda$) for both Baseline (non-DP) and DP models.} \label{fig:cifar10_l2}
\end{figure}

Additionally, Table \ref{fig:cifar10_l2} and Figure \ref{fig:cifar10_l2} illustrate the effects of applying $L_2$ regularization to the models. As the regularization strength parameter $\lambda$ increases, the non-DP model exhibits a clear decline in both validation accuracy and attacker advantage. However, the DP model with $L_2$ regularization demonstrates a more resilient behavior, maintaining or even improving performance at higher $\lambda$ values. Notably, starting from $\lambda=0.001$, the non-DP model achieves the highest validation accuracy of 73.3\% while also maintaining a lower attacker advantage of 5.76 compared to the non-DP model's of 9.25. This suggests in this experiment the model does not suffer as much overfitting as the non-DP model. As $\lambda$ increases to $0.003$, the validation accuracy decreases to 67.8\% but remains higher than that of the DP model's 59.7\%. Finally, at $\lambda=0.005$, it continues to outperform the DP model in terms of validation accuracy with 64.0\% and also exhibits the lowest attacker advantage among all models of 1.21. Additionally, the DP model, when trained with $L_2$ regularization, achieves a validation accuracy of approximately 58.36 and an attacker advantage of 1.58, regardless of the regularization strength parameter. It is evident that the non-DP model, when trained with regularization, doesn't achieve a comparable level of performance. Even though it may exhibit a lower attacker advantage, its validation accuracy is worse than other models.

\subsection{Text Classification Task}

The dataset employed for this task was an artificially enhanced version of the Toxic Tweets Dataset \cite{kaggleToxicTweets}.
The original dataset consisted of 54,313 tweets, where each tweet is labeled as either toxic or non-toxic with each class containing 24153 and 32592 posts respectively. The dataset was enhanced by adding a new column, called "vulnerability", which indicates whether the author of the tweet belongs to a vulnerable group or not. The vulnerable group is defined as a group of people who are more likely to be affected by bias in AI models. Vulnerable groups include, but are not limited to, elderly people, young adults, racial minorities, and people with disabilities, as identified in the citizen study conducted in the pilot cities of Martin and Brasov for the purposes of the ITHACA project \cite{ithaca}.

The vulnerability was determined via a random process guided by the results of the citizen study, where the probability of a tweet being authored by a member of a minority group was 20\%. To test the effectiveness of the method proposed in this work, an artificial bias was injected into the dataset, where posts originating from the vulnerable groups were more likely to be labeled as toxic, with a probability of 70\%. This signifies that for a randomly selected post, there is a 70\% chance that it will be labeled as toxic if it was authored by a member of a vulnerable group, and a 30\% chance that it will be labeled as toxic if it was authored by a non-vulnerable group.

This artificial bias was introduced to simulate the real-world scenario where AI models are trained on biased datasets, which has been shown to lead in biased predictions made by the model \cite{pmlr-v108-jiang20a}.

Our baseline non-DP model is a Natural Language Processing Model for detecting the toxicity of text excerpts contained in the dataset. The model consisted of a text vectorization pre-processing layer, followed by an embedding layer responsible for converting the input data into a continuous dense vector of fixed size, where values capture the semantic meaning or relationship between the inputs, a Global Average Pooling layer, and a Dense layer with a sigmoid activation function. The maximum vocabulary length was set to $10^4$ and the maximum output sequence length for the text vectorizer was set to 15 words per post. The output dimension of the embedding layer was set to 128, and the output dimension of the Dense layer was set to 1, as the model was trained as a binary classifier, with the goal of predicting whether a given text excerpt is toxic or non-toxic.

The non-DP models were trained using the Adam optimizer with a learning rate of 0.001 for 100 epochs and a batch size of 1024. For the DP models, we used the DP-Adam optimizer from TensorFlow Privacy, setting the noise multiplier to 1.1 and the clipping norm to 1.0.

\begin{figure}[ht!]
\centering
\includegraphics[width=\textwidth]{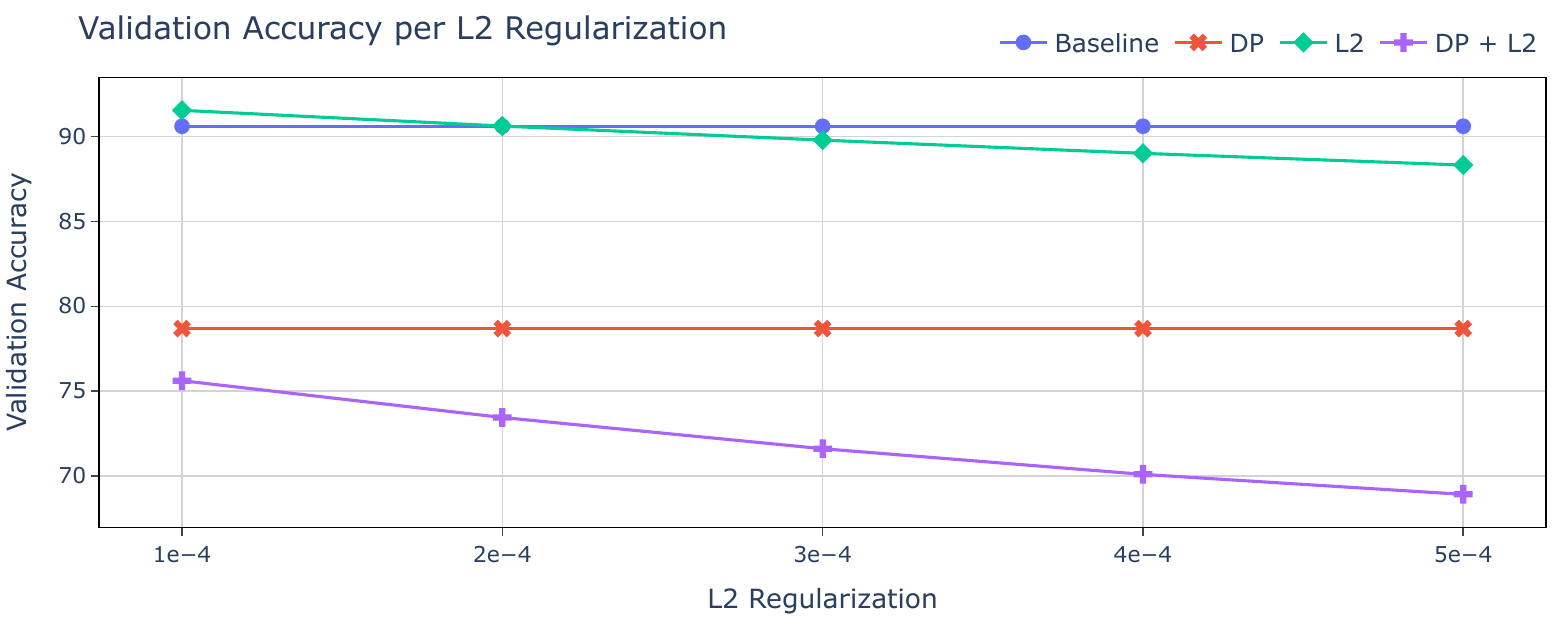}\\
\includegraphics[width=\textwidth]{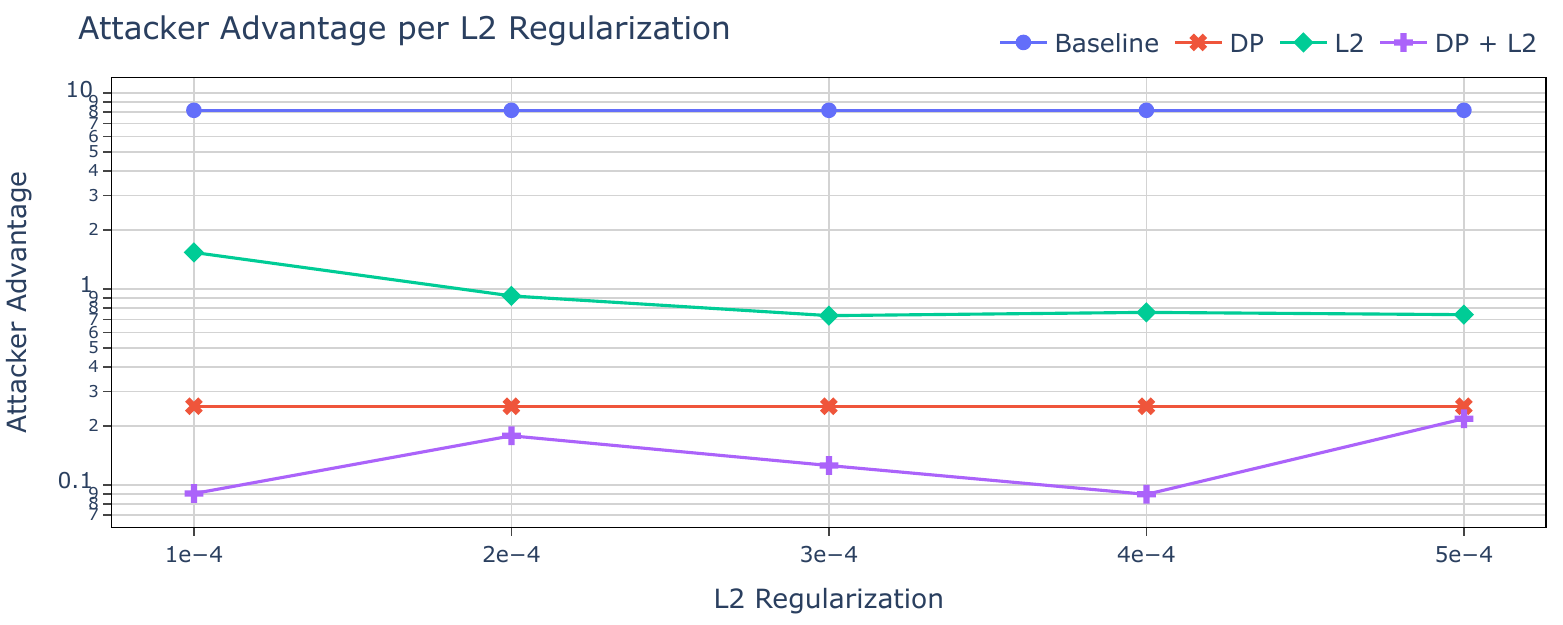}\\
\caption{Effect of $L_2$ regularization on the text classification task. The top plot shows the validation accuracy across different $L_2$ regularization strengths ($\lambda$) for both Baseline (non-DP) and DP models. The bottom plot illustrates the corresponding attacker advantage for each model.}  \label{fig:txt}
\end{figure}

The experimental results are presented in Table \ref{table:txt} and visualized in Figure \ref{fig:txt}. The table reports the average and standard deviation of training accuracy, validation accuracy, and attacker advantage over five runs for both Baseline (non-DP) and DP models, across varying $L_2$ regularization strengths ($\lambda$) ranging from $1\times10^{-4}$ to $5\times10^{-4}$.

\begin{table}[ht!]
\caption{Performance metrics for the text classification task with varying $L_2$ regularization strengths ($\lambda$). The table presents the average and standard deviation of training accuracy, validation accuracy, and attacker advantage over five runs for both Baseline (non-DP) and DP models.}\label{table:txt}
\centering
\small
\setlength{\tabcolsep}{3pt}
\begin{tabular}{l|cccccc}
\hline
\hline
 & \multicolumn{6}{|c}{Train Accuracy} \\ 
\hline
\hline
Baseline & $97.9 \pm .001$ & $92.7 \pm .001$ & $91.1 \pm .001$ & $90.0 \pm .001$ & $89.3 \pm .001$ & $88.7 \pm .002$\\
DP       & $78.4 \pm .001$ & $75.7 \pm .005$ & $73.3 \pm .006$ & $71.5 \pm .011$ & $70.0 \pm .020$ & $68.7 \pm .016$\\
\hline
\hline
 & \multicolumn{6}{|c}{Validation Accuracy} \\
\hline
\hline
Baseline & $90.6 \pm .014$ & $91.5 \pm .004$ & $90.6 \pm .003$ & $89.7 \pm .003$ & $89.0 \pm .004$ & $88.3 \pm .003$\\
DP       & $78.6 \pm .035$ & $75.6 \pm .017$ & $73.4 \pm .022$ & $71.6 \pm .009$ & $70.0 \pm .044$ & $68.9 \pm .030$\\
\hline
\hline
 & \multicolumn{6}{|c}{Attacker Advantage} \\
\hline
\hline
Baseline & $8.16 \pm .006$ & $1.53 \pm .004$ & $0.92 \pm .001$ & $0.73 \pm .003$ & $0.75 \pm .003$ & $0.73 \pm .003$\\
DP       & $0.25 \pm .007$ & $0.09 \pm .008$ & $0.17 \pm .013$ & $0.12 \pm .022$ & $0.09 \pm .025$ & $0.21 \pm .018$\\
\hline
\hline
$\lambda$ & $0$ & $.0001$ & $.0002$ & $.0003$ & $.0004$ & $.0005$\\
\hline
\end{tabular}
\end{table}

From Table \ref{table:txt}, we observe that the Baseline model achieves significantly higher training accuracy compared to the DP model across all $\lambda$ values. At $\lambda=0$, the Baseline model reaches a training accuracy of 97.99\%, while the DP model attains 78.44\%. As $\lambda$ increases, both models exhibit a decrease in training accuracy, with the Baseline model showing a more pronounced decline due to the stronger regularization effect. In terms of validation accuracy, the Baseline model starts at 90.61\% for $\lambda=0$ and decreases slightly to 88.32\% at $\lambda=0.005$. Interestingly, at $\lambda=0.001$, the Baseline model's validation accuracy slightly improves to 91.55\%, suggesting that mild regularization can enhance generalization by preventing overfitting. The DP model's validation accuracy closely follows its training accuracy, starting at 78.68\% and decreasing to 68.91\% as $\lambda$ increases.

Regarding the attacker advantage, the Baseline model shows a substantial decrease as $\lambda$ increases. At $\lambda=0$, the attacker advantage is 8.16, which significantly reduces to approximately 0.73 at $\lambda=0.005$. This indicates that increasing the $L_2$ regularization strength effectively enhances the model's privacy by reducing its vulnerability to Membership Inference Attacks. The DP model consistently maintains a very low attacker advantage across all $\lambda$ values, ranging from 0.25 to 0.09, demonstrating the strong privacy protection inherent in DP training.

Figure \ref{fig:txt} illustrates the impact of $L_2$ regularization on validation accuracy and attacker advantage for both Baseline and DP models. The top plot shows that the validation accuracy of the Baseline model remains relatively stable with increasing $\lambda$, with a slight improvement at $\lambda=0.001$, before gradually decreasing at higher $\lambda$ values. The DP model experiences a consistent decline in validation accuracy as $\lambda$ increases. The bottom plot highlights the significant reduction in attacker advantage for the Baseline model as $\lambda$ increases, whereas the DP model maintains a consistently low attacker advantage across all regularization strengths. 

These results suggest that applying $L_2$ regularization to the Baseline model not only improves its generalization performance but also significantly enhances its privacy by reducing the attacker's advantage. At $\lambda=0.001$, the Baseline model achieves its highest validation accuracy of 91.55\% while substantially lowering the attacker advantage from 8.16 to 1.53. This indicates that even mild regularization can have a noticeable effect on both accuracy and privacy. For higher values of $\lambda$, the Baseline model continues to reduce the attacker advantage, approaching values similar to the DP model, while maintaining higher validation accuracy. At $\lambda=0.005$, the Baseline model's attacker advantage decreases to 0.73, closely aligning with the DP model's range. However, the validation accuracy remains significantly higher than that of the DP model, demonstrating that $L_2$ regularization can be an effective strategy for balancing privacy and performance. The DP model, while offering strong privacy protection across all $\lambda$ values, exhibits lower validation accuracy compared to the Baseline model. This performance gap highlights the trade-offs associated with DP training, where the addition of noise to achieve privacy can affect model accuracy.

\subsection{Overfit Evaluation}
These results underscore the potential of $L_2$ regularization to provide better privacy protections, as it both reduces the attacker advantage and sustains strong validation performance, in contrast to the DP-optimized model. Furthermore, the strength parameter $\lambda$ can be seen as a trade-off factor between model accuracy and privacy. Lower values of $\lambda$ tend to result in higher accuracy but with reduced privacy protection, as reflected in higher attacker advantage. Conversely, increasing $\lambda$ enhances privacy by lowering the attacker advantage, but at the cost of reduced model accuracy. This trade-off can be tuned, for the specific use case, to strike a balance between maintaining strong validation performance and minimizing the model's vulnerability to attacks.

\begin{figure}[ht!]
\centering
\includegraphics[width=\textwidth]{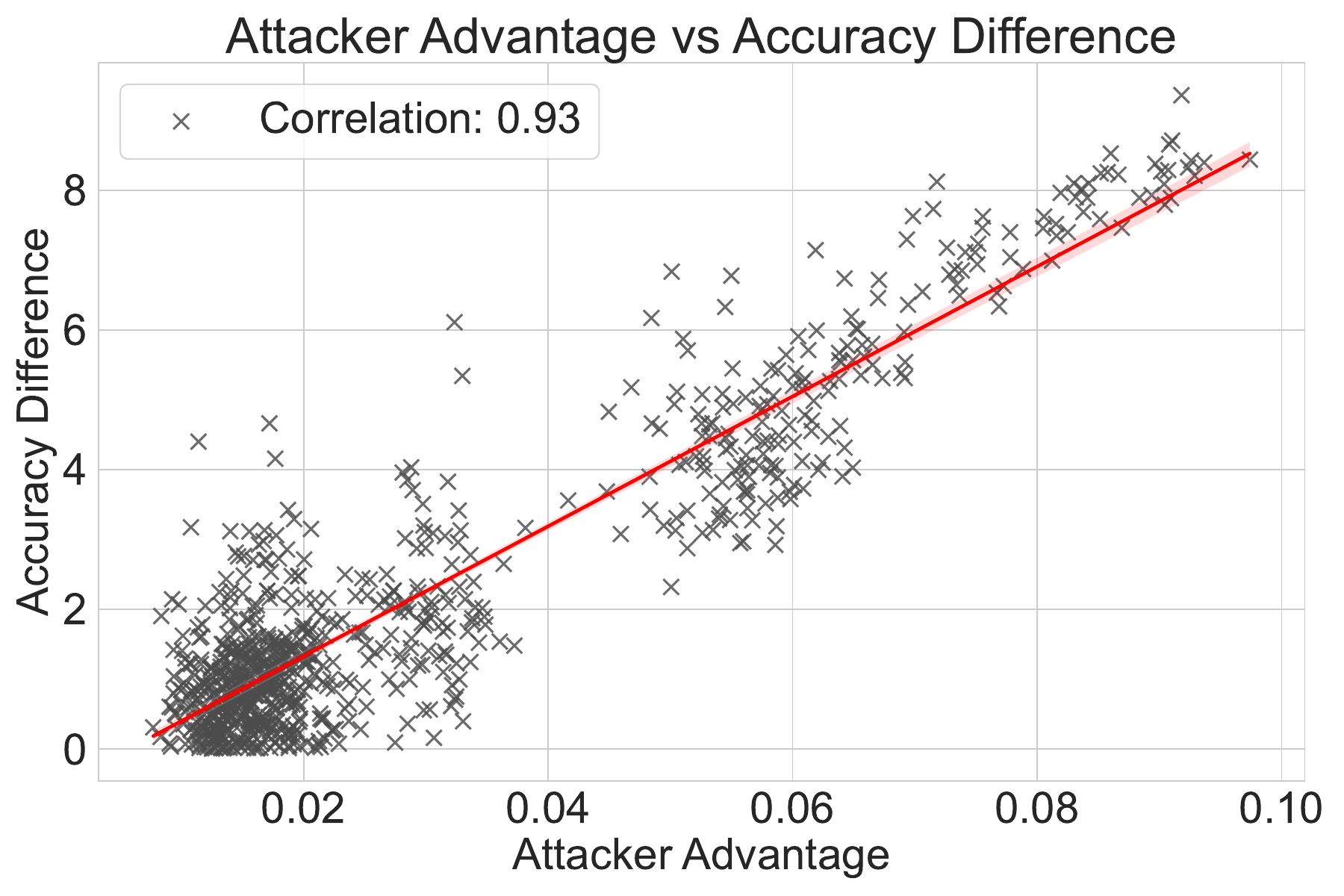}
\caption{Correlation between the accuracy difference (training accuracy minus validation accuracy) and attacker advantage. The plot demonstrates a strong positive correlation (correlation coefficient of 0.93).} \label{fig:acc_atk}
\end{figure}

Finally, we analyzed the relationship between the difference in training and evaluation accuracy (accuracy difference) and the attacker advantage. To visualize this relationship, we plotted the accuracy difference on the y-axis and the attacker advantage on the x-axis. The resulting plot, presented in Figure \ref{fig:acc_atk}, demonstrates that as the accuracy difference increases, the attacker's advantage also increases. Specifically, we observed a correlation coefficient of 0.93, indicating a strong positive correlation between these two metrics.

These findings indicate that the gap between training and evaluation accuracy is a significant factor contributing to the vulnerability of models to MIAs. A higher accuracy difference suggests that the model has overfitted to the training data, capturing noise and specific patterns that do not generalize well to unseen data. This overfitting makes it easier for an attacker to distinguish whether a particular data point was part of the training set, thereby increasing the attacker's advantage. Maintaining a lower gap between training and evaluation accuracy is therefore crucial for reducing vulnerability to MIAs. Techniques that mitigate overfitting, such as $L_2$ regularization, and therefore dropout, early stopping, should be prioritized. Our experiments demonstrate that applying $L_2$ regularization not only enhances the model's generalization capabilities but also effectively reduces the attacker's advantage, sometimes outperforming models trained with differential privacy in terms of both validation accuracy and privacy protection.

\section{Conclusion} \label{sec:conclusions}
We have investigated the effectiveness of $L_2$ regularization in enhancing privacy protection against MIAs in DL models. Our approach was evaluated on benchmark datasets, including MNIST \cite{mnist}, CIFAR-10 \cite{cifar10}, and an augmented version of the Toxic Tweets Dataset \cite{kaggleToxicTweets}, using various neural network architectures.

Overall, the analysis of the experimental results demonstrates that incorporating $L_2$ regularization into model training can reduce the attacker's advantage while maintaining or even improving model accuracy compared to models trained with DP. Specifically, our findings indicate that $L_2$ regularization effectively mitigates overfitting, which is a key factor contributing to the vulnerability of models to MIAs. Models trained with $L_2$ regularization not only achieved higher validation accuracies but also exhibited enhanced privacy protection.

Despite the improvements offered by $L_2$ regularization, some limitations still exist. The effectiveness of $L_2$ regularization in providing privacy protection is closely tied to its ability to prevent overfitting. In scenarios where models are inherently prone to overfitting due to complex data distributions or limited training data, $L_2$ regularization alone may not be sufficient to ensure robust privacy. Additionally, while $L_2$ regularization can reduce the attacker's advantage, it does not provide formal privacy guarantees like differential privacy does.

To further enhance privacy protection, future research could explore combining $L_2$ regularization with other regularization techniques. Moreover, integrating $L_2$ regularization with formal privacy-preserving mechanisms like DP could offer a balanced approach that leverages the strengths of both methods. Future work could also focus on extending the evaluation to more complex datasets and models to generalize the findings and develop robust strategies for privacy-preserving machine learning.

\section*{Data Availability}
All data underlying the analyses are available as part of the article or as referenced external data sources and no additional source data are required.

\section*{Declaration of competing interest}
The authors declare that they have no known competing financial interests or personal relationships that could have appeared to influence the work reported in this paper.

\bibliographystyle{splncs04}
\bibliography{mybibliography}




% LNCS Homepage, \url{http://www.springer.com/lncs}. Last accessed 4
% \end{thebibliography}
\end{document}